%% file: paper.tex
\documentclass[runningheads]{llncs}

\usepackage[T1]{fontenc}
\usepackage{graphicx}
\usepackage[utf8]{inputenc} 
\usepackage[T1]{fontenc}    
\usepackage{hyperref}       
\usepackage{url}            
\usepackage{booktabs}       
\usepackage{amsfonts}       
\usepackage{amsmath}        
\usepackage{amsfonts}       
\usepackage{nicefrac}       
\usepackage{microtype}      
\usepackage{xcolor}         

\usepackage{amsfonts}
\usepackage{commath}
\usepackage{cleveref}
\usepackage{placeins}
\usepackage{booktabs}
\usepackage{enumitem}
\usepackage{comment}   
\makeatletter
\newcommand{\printfnsymbol}[1]{%
  \textsuperscript{\@fnsymbol{#1}}%
}
\makeatother

\titlerunning{Misconception Diagnosis From Student-Tutor Dialogue}
\begin{document}
\title{Misconception Diagnosis From Student-Tutor Dialogue: Generate, Retrieve, Rerank}
%
%
\author{Joshua Mitton\textsuperscript{1}\thanks{Equal contribution}\and
 Prarthana Bhattacharyya\textsuperscript{1}\printfnsymbol{1} \and
 Digory Smith\inst{1} \and
 Thomas Christie\inst{2} \and
 Ralph Abboud\inst{3} \and
 Simon Woodhead\inst{1}}
\authorrunning{J. Mitton et al.}
\institute{Eedi \and
 Renaissance Philanthropy \and
 Learning Engineering Virtual Institute\\
}
%
%
%

\maketitle              
\begin{abstract}
Timely and accurate identification of student misconceptions is key to improving learning outcomes and pre-empting the compounding of student errors. However, this task is highly dependent on the effort and intuition of the teacher. In this work, we present a novel approach for detecting misconceptions from student-tutor dialogues using large language models (LLMs). First, we use a fine-tuned LLM to generate plausible misconceptions, and then retrieve the most promising candidates among these using embedding similarity with the input dialogue. These  candidates are then assessed and re-ranked by another fine-tuned LLM to improve misconception relevance. Empirically, we evaluate our system on real dialogues from an educational tutoring platform. We consider multiple base LLM models including LLaMA, Qwen and Claude on zero-shot and fine-tuned settings. We find that our approach improves predictive performance over baseline models and that fine-tuning improves both generated misconception quality and can outperform larger closed-source models. Finally, we conduct ablation studies to both validate the importance of our generation and reranking steps on misconception generation quality. 

\end{abstract}

\begin{figure}[htb!]
    \centering
    \includegraphics[width=\textwidth]{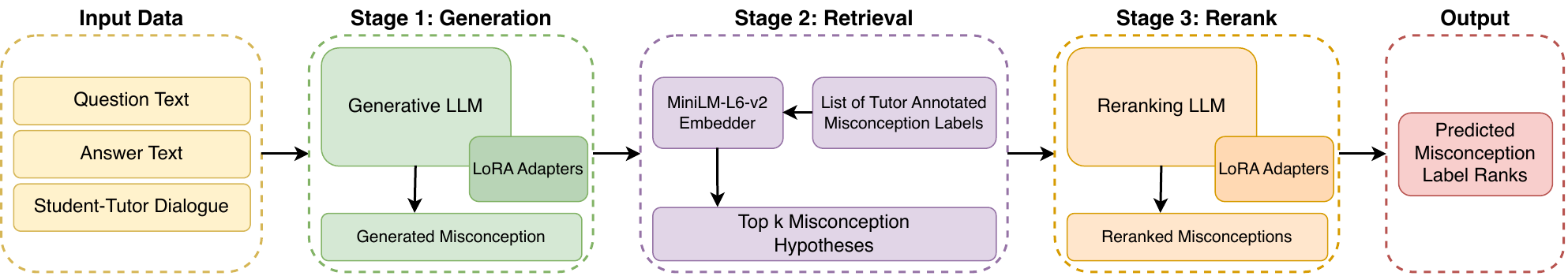}
    \caption{Model architecture diagram showing the three stages of our model. The LoRA adapters, shown with a darker background, are the trainable parts of the model. Stage 1 is a fine-tuned LLM for generating misconceptions. Stage 2 is a retrieval step using an embedding model and cosine similarity to predict the top k misconceptions. Stage 3 is a fine-tuned LLM for reranking the top k misconceptions.}
    \label{fig:model}
\end{figure}

\section{Introduction}
\input{chapters/introduction}

\section{Background: Misconception Diagnosis}
\input{chapters/background_related_work}

\section{Experimental Setup}
\input{chapters/experimental_setup}

\section{Models}

\subsection{Proposed Model}
\input{chapters/1_proposed_model}

\subsection{Baselines}
\input{chapters/2_baseline_models}

\subsection{Base LLM Models}
\input{chapters/3_closedS_openS_models}

\subsection{Fine-tuning LoRA}
\input{chapters/4_finetuning_LoRA}

\vspace{-0.3cm}
\section{Results}
\input{chapters/results}

\section{Discussion}
\input{chapters/discussion}

\section{Conclusion}
We present a novel model pipeline comprising of generation-retrieval-reranking that outperforms baselines in the task of misconception diagnosis from diagloue. We develop a parameter efficient LoRA fine-tuning approach that improves the generated misconception style and can outperform closed source models an order of magnitude larger. We perform extensive ablations of each model component and benchmarking analysis to demonstrate the importance of each proposed component in our framework.

\begin{credits}
\subsubsection{\discintname}
The authors have no competing interests to declare that
are relevant to the content of this article.
\end{credits}
\vspace{-0.1cm}
%
%
%
\bibliographystyle{splncs04}
\bibliography{references}
%
\newpage
\appendix
\section*{Appendix}
\input{chapters/appendix}
\end{document}

%% file: chapters/introduction.tex
Students often hold misconceptions, which are systematic misunderstandings of a given concept leading to predictable error patterns. For example, a student can excessively move the decimal point during division by 10, leading to answers consistently in the wrong order of magnitude. Misconceptions, left unchecked, can compound and exacerbate a student's difficulties during their learning journey. It is therefore imperative to identify misconceptions early and intervene effectively. 

Despite progress with large language models (LLMs) in recent years, identifying student misconceptions remains a challenging task for many reasons. First, a misconception is ambiguous to define, with the granularity and scope of a misconception often making detection an ad-hoc and intuition-driven process. Second, misconceptions often overlap, particularly when being assessed against a shorter history of student activity \cite{wang2020diagnostic}. For instance, the earlier decimal point misconception can lead to similar erroneous answers as another misconception: Multiplying when asked to divide by powers of 10. Early recognition of these misconception patterns allows educators to provide targeted interventions and improve learning outcomes. Therefore, diagnosing misconceptions is of importance for intelligent tutoring systems and educational learning platforms. 

One convenient way for educational platforms to frame teacher interventions is through multiple choice questions, with distractors tied to clearly defined misconceptions. With these questions, knowledge tracing models \cite{AKT} can be trained on student response data to predict which distractor a student will select given their response history, thereby indicating their current misconceptions. However, this setting provides a limited training signal (only answer correctness and the question and answer texts), and poor surrounding context regarding students' meta-behaviours, e.g., slips, engagement, and guessing. Consequently, these models often peak at around 65-70\% accuracy. 

\begin{figure}[t]
    \centering
    \includegraphics[width=0.8\textwidth]{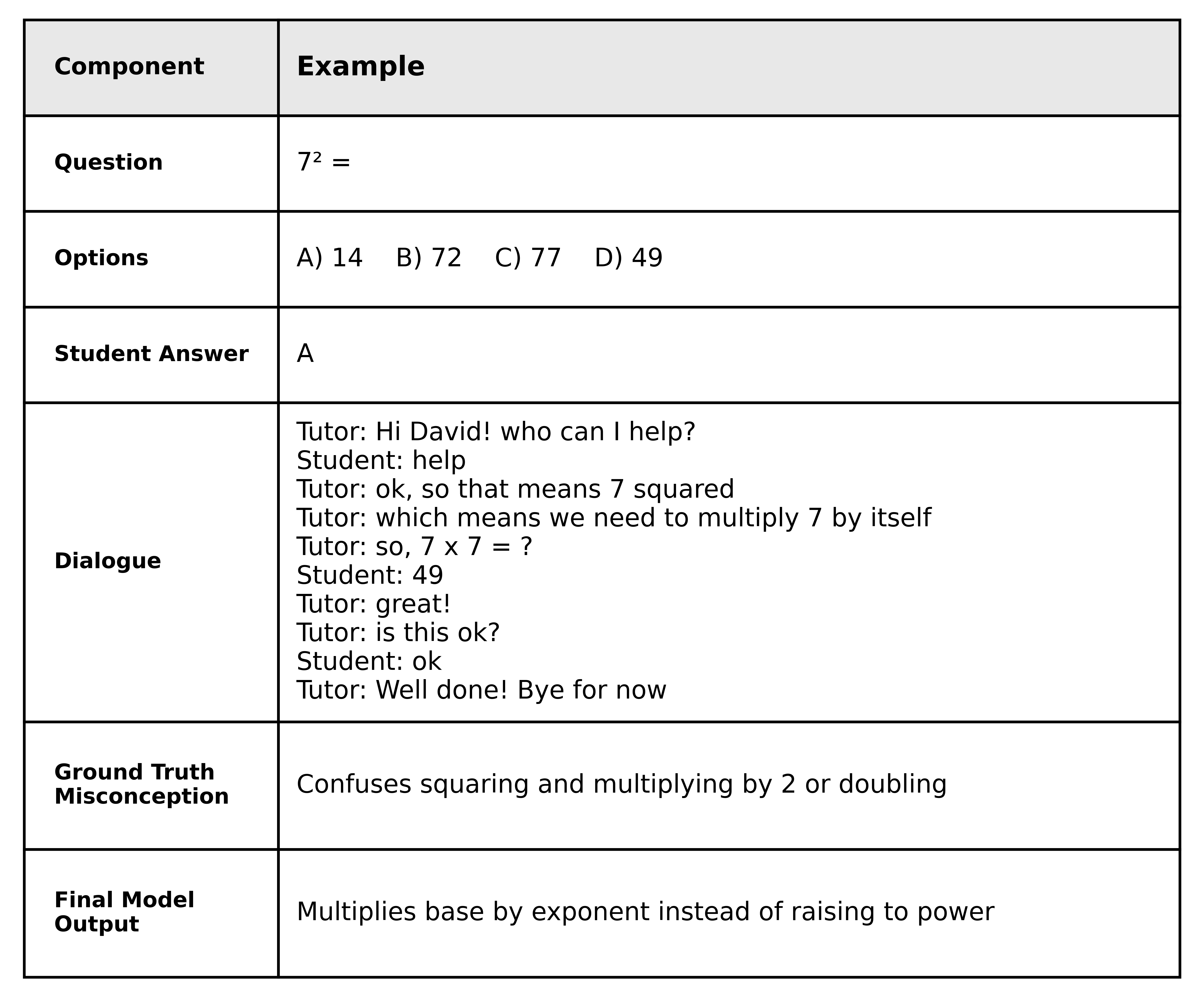}
    \vspace{-0.5cm}
    \caption{An example data point from our task setting. The input to our proposed approach is a multiple choice question in maths, paired with the student's chosen answer, the corresponding tutor-student dialogue, and the tutor's misconception diagnosis as the ground truth.} 
    \label{fig:intro_pic}
    \vspace{-0.3cm}
\end{figure}

In this work, we investigate whether a machine learning model can instead identify student misconceptions from student-tutor dialogues, where the tutor probes the student's understanding of a question and their chosen answer. This richer context has the potential to improve misconception diagnosis capability, but in turn introduces the added complexity of mapping student dialogue, to generalizable, yet specific, tutor defined misconception labels. An example of a misconception tied to a multiple choice mathematics question is given in \Cref{fig:intro_pic}.


Our approach, as illustrated in \Cref{fig:model}, introduces a generation-retrieval-reranking pipeline. 
Rather than prompting an LLM to directly predict a misconception, we use an LLM to generate a misconception hypothesis, then match in embedding space against tutor-defined labels, followed by LLM-based reranking. 

\noindent Our main contributions are as follows.
\begin{enumerate}[nosep]
    \item A novel model that improves over baselines in the task of misconception diagnosis from dialogue.
    \item A parameter efficient LoRA fine-tuning approach that improves the generated misconception style and outperforms significantly larger closed source models.
    \item An extensive ablation and benchmarking analysis that establishes the importance of each of the proposed components of our approach.
\end{enumerate}

%% file: chapters/background_related_work.tex
\subsection{Related Work} In the literature, misconception diagnosis has been addressed by creating tutor-annotated ground-truth misconception labels of common misunderstandings and manually mapping student errors to predefined categories \cite{brown1978diagnostic,vanlehn1990mind,feldman2018automatic}. Recent benchmark-style datasets have redefined this setting as natural language processing (NLP) formulations, such as predicting the affinity between distractors and ground-truth misconception labels \cite{eedi-mining-misconceptions-in-mathematics}. Contemporary work explores both generative and retrieval approaches for grounding noisy student artifacts to canonical misconception descriptors, as well as hybrid methods that combine language models with traditional knowledge tracing systems \cite{scarlatos2025exploring,MIRAGE,ross2025learning}.

\subsection{Generative LLMs for Misconception Diagnosis} While LLMs show promise in this domain, research reveals important limitations in their current capabilities. LLMs have been shown to infer student misconceptions and adapt teaching strategies better than simple baselines but worse than methods that explicitly model misconceptions \cite{ross2024toward}. In addition, LLMs struggle to identify incorrect reasoning that contains misconceptions more than they do correct reasoning \cite{sonkar2024malalgoqa}. Combining language models with knowledge tracing has been demonstrated to lead to better estimates of student knowledge states than knowledge tracing-only methods in dialogue settings \cite{scarlatos2025exploring}. The recently proposed MISTAKE framework \cite{ross2025learning} explicitly targets misconception modelling by generating synthetic reasoning traces containing realistic errors, demonstrating gains on both student simulation and misconception inference tasks. These directions motivate using an LLM not only to choose a misconception label from predefined options, but also to produce an interpretable intermediate hypothesis about the student's underlying misconception.
\vspace{-0.4cm}

\subsection{Misconception Diagnosis as Retrieval} An alternative approach frames misconception identification as semantic retrieval. Student responses and misconception labels are embedded into a shared space, and the nearest label by cosine similarity is selected. Methods like Sentence-BERT \cite{reimers2019sentence} and DPR \cite{karpukhin2020dense} enable efficient similarity search and naturally scale to large misconception taxonomies. However, retrieval degrades when student responses are short or symbolic. Recent embedding models trained with contrastive objectives \cite{wang2024multilingual} and instruction-aware models \cite{su2023one} have improved semantic matching on general benchmarks \cite{muennighoff2023mteb}. Despite this, evaluation frameworks reveal substantial performance variation across domains. This suggests general-purpose embeddings may not transfer optimally to educational applications where the gap between student error language and expert misconception descriptions remains a key challenge.
\vspace{-0.4cm}


\subsection{Datasets} A key challenge for misconception detection is the scarcity of high-quality data. This is particularly true for expert-annotated examples of real student misconceptions. While datasets like DrawEduMath contain students' handwritten solutions annotated by expert teachers \cite{baral2025drawedumath}, they lack standardized annotations of higher-level misconceptions. Similarly, the MalAlgoQA dataset \cite{sonkar2024malalgoqa} contains math problems with associated incorrect answers and rationales, but these rationales remain problem-specific. The Eedi dataset \cite{eedi-mining-misconceptions-in-mathematics} represents one of a few resources that contains natural student data with annotations of student misconceptions. Despite this, the misconception labels are very general, lacking specific detail, and the suggested dataset splits are split randomly, causing the misconception labels to be shared between the train and test sets, which does not allow model generalisation to be assessed.

%% file: chapters/experimental_setup.tex
\subsection{Predictive Task}
 The task we are seeking to address in this paper is to be able to predict student misconceptions. The data available to a model consists of the question the student answered, the answer the student chose, and a dialogue the student had with a tutor. The task of the model is to predict the ground truth misconception provided by a tutor. An example of the input data for the task is given in \Cref{fig:intro_pic}. 

\subsection{Dataset}
We collect a new tutor-student dialogue dataset for this task. The true labels of the misconception the student has were provided by human tutors who labelled how likely it is that the student has a specific misconception. The tutors labelled the likelihood that the student has a misconception from possible labels of \{0, 25, 50, 75, 100\}, where 0 indicates that the student does not have the misconception and 100 indicates that they definitely have the misconception. For this paper, we consider the task of a model predicting the misconception a student has from the ground-truth misconception labels, and therefore only retain data where the tutor labelled the likelihood of the student having a misconception as 75 or 100. This gives us 922 student-tutor conversations as data points, which we have collected from an online educational learning platform. The total number of unique ground-truth misconception labels in our dataset is 546. This means the majority of misconceptions only occur once, producing a difficult prediction task.

When splitting the dataset into a training, validation, and testing set we chose to split the dataset on unique misconception labels. We chose to split the datasets by misconception rather than a simple random split to ensure that misconception targets are unique between each dataset. This split has an advantage over splitting randomly because it ensures that when testing a trained model the model's ability to generalise to new unseen misconceptions is assessed. We use 70\% of the data for training, 10\% for validation and 20\% for testing.

\subsection{Experimental Protocol}
For each metric, $N$ is the number of data points, $r_{i}$ is the predicted rank of the true misconception for the $i^{\mathrm{th}}$ data point, and $k$ is the cut off rank above which $0$ is scored. The first metric we consider to assess the model performance is MAP@$k$ (Mean Average Precision at $k$). This metric evaluates how good a model is at sorting relevant items at the top of a ranking list of retrieved items. The equation we use for MAP@$k$ is:
\begin{equation}
    \mathrm{MAP@}k = \frac{1}{N} \sum^{N}_{i=1} \mathbb{I}\left( r_{i} \leq k \right) \frac{1}{r_{i}}.
    \label{eq:map}
\end{equation}
Note that this is a simplified MAP@k equation due to there only being one relevant misconception for each input.

The next metric we use to consider is Normalized Discounted Cumulative Gain (NDCG). NDCG also evaluates how good a model is at sorting relevant items, but accounts for the entire list and takes the logarithm of the rank proving a slower growing denominator with rank. This results in NDCG penalising the model less for not scoring rank 1. The equation we use for NCDG is:
\begin{equation}
    \mathrm{NDCG} = \frac{1}{N} \sum^{N}_{i=1} \frac{1}{\mathrm{log}_{2}\left( r_{i}+1 \right)}.
    \label{eq:ndcg}
\end{equation}
This is the standard NDCG equation simplified due to there only being one single ground truth misconception for each input dialogue.

We also consider the recall@k to assess the ability of the model. The recall@$k$ shows how often a model predicts the correct misconception in the top $k$. The equation we use for recall@$k$ is:
\begin{equation}
    \mathrm{Recall@}k = \frac{1}{N} \sum^{N}_{i=1} \mathbb{I}\left( r_{i} \leq k \right).
    \label{eq:recall}
\end{equation}

The other metrics we consider are cosine similarity of the generated misconception hypothesis with the true misconception label, and mean and median retrieval rank, described in the appendix.


%% file: chapters/1_proposed_model.tex
Our proposed model for the task of diagnosing misconceptions from student-tutor dialogue is a three stage model comprised of a generation stage, an embedding retrieval stage, and a reranking stage. A diagrammatic form of our model is presented in \Cref{fig:model}. Our model takes as input the question text, the answer text, and a dialogue between the student and a tutor. Stage 1 takes the input data and uses an LLM to generate a plausible hypothesis for a misconception that the student could have. Stage 1 also has parameter efficient fine-tuned LoRA layers, which improves the style of the generated misconceptions. Stage 2 conducts a fast and efficient retrieval using the off-the-shelf MiniLM-L6-v2 \cite{wang2020minilm} embedding model to embed the generated misconception and each ground-truth misconception label. The top-k misconceptions are retrieved by ranking based on cosine similarity between generated and ground-truth misconception label embeddings. Stage 3 uses an LLM to rerank the top $k$ misconception labels. This stage is used to better capture the semantics of the misconceptions. Stage 3 also has parameter efficient fine-tuned LoRA layers that allow the model in this stage to become specialised in the task of ranking misconceptions. At the end of stage 3 the highest ranked misconception label is considered the predicted misconception. LoRA layers for stage 1 and 3 are trained separately.

%% file: chapters/2_baseline_models.tex
\label{baselines}
We compare against three baselines to isolate the contribution of our generation-retrieval-reranking approach. Each baseline tests a different aspect of the pipeline.

\textit{Direct Embedding Matching (No LLM Generation).} This baseline skips the LLM generation step entirely. It directly embeds student dialogue text using MiniLM-L6-v2 \cite{wang2020minilm}. The embeddings are then matched against misconception labels via cosine similarity. This tests whether the LLM generation adds value over direct semantic matching. It also reveals whether direct embedding of conversational turns suffices for misconception identification. 

\textit{Zero-shot LLM Classification.} This baseline presents Claude Sonnet 4.5 with the student dialogue and the complete list of 546 unique misconception labels. The model is prompted to directly rank the top-10 most likely misconceptions in a single inference step. The prompt includes the math question, answer options, and the student's selected choice alongside the dialogue. This tests whether decomposing the task into generation and retrieval is necessary, or whether a capable LLM can navigate large label spaces in a single inference pass.

\textit{TF-IDF Keyword Matching.} This non-neural baseline represents dialogues and misconception labels using TF-IDF \cite{sparck1972statistical} vectors. Rankings are determined by cosine similarity measuring keyword overlap between student dialogues and misconception descriptions. This tests whether simple lexical overlap suffices for misconception retrieval. 

%% file: chapters/3_closedS_openS_models.tex
The generation component requires a LLM model that can interpret student dialogue and generate the underlying misconceptions. We evaluated both closed-source and open-source models for this task.

Closed-source LLMs like Claude Sonnet 4.5 \cite{anthropic2025claudeSonnet45SystemCard} offer strong instruction-following and reasoning capabilities. Because they're pre-trained on massive datasets that likely include educational content, they work well for zero-shot misconception diagnosis. However, API costs add up with each request, and sending student data to external servers can raise privacy concerns. We also evaluated open-source LLMs like Llama 3.2 3B Instruct \cite{grattafiori2024llama3herd} and Qwen 2.5 7B Instruct \cite{qwen2025qwen25technicalreport}. Both of these models can be deployed locally. This gives better data privacy and cheaper hosting costs. Smaller models like Llama 3.2 3B are more efficient, while larger models like Qwen 2.5 7B tend to show better reasoning and language understanding. Closed-source models are often much larger, though their exact sizes aren't publicly disclosed. In the end, choosing between model types involves weighing trade-offs among performance, deployment flexibility, computational requirements, cost, and privacy.

%% file: chapters/4_finetuning_LoRA.tex
While prompt engineering works well for larger models, smaller open-source models still struggle to articulate misconceptions from dialogue. To address this, we use Low-Rank Adaptation (LoRA) \cite{hu2022lora} to fine-tune both LLaMA 3.2 3B Instruct and Qwen 2.5 7B Instruct on a 70\% split our annotated dataset. Fine-tuning teaches the models to produce concise, retrieval-friendly misconception labels. The base models are verbose and don't do this reliably out of the box. LoRA keeps only about 0.5\% of parameters trainable, making the process computationally efficient. After fine-tuning, we observe improved cosine similarity between model-generated descriptions and expert-authored misconception labels in the embedding space. This suggests the LoRA models learn to produce descriptions that are stylistically closer to tutor annotations.

%% file: chapters/results.tex
\subsection{Baseline Models}
The three baseline models described in \Cref{baselines} establish lower bounds for (1) skipping the generative hypothesis prediction layer, (2) direct LLM classification over large label spaces without generation or retrieval, and (3) pure lexical matching without semantic understanding.


Figures~\ref{fig_map} and \ref{fig_ndcg} show that keyword matching outperforms a zero-shot LLM classification and directly using an embedding model on metrics that prioritise top-1 prediction. Figure~\ref{fig_recall} shows that direct embedding is stronger than keyword matching and LLM classification when the focus is getting the most predictions in the top $k$ rank. 
\par Overall, the zero-shot LLM classification baseline scores poorly, likely due to struggling to handle the large context with 546 unique misconceptions, while the keyword matching and direct embedding methods provide baseline values a custom model is required to outperform.
\vspace{-0.3cm}
\subsection{Base LLM Model}
In this section, we compare the ability of the base LLM model when running the three stages of generation, embedding and reranking. Figures~\ref{fig_map}, \ref{fig_ndcg}, \ref{fig_cosinesim}, \ref{fig_recall} and \ref{fig_rank}  show that the Llama 3.2 3B Instruct \cite{grattafiori2024llama3herd} model is significantly outperformed by Claude Sonnet 4.5 \cite{anthropic2025claudeSonnet45SystemCard} and Qwen 2.5 7B Instruct \cite{qwen2025qwen25technicalreport}. This is due to Llama 3.2 3B struggling to accomplish the task of reranking, significantly hurting its overall performance. Impressively, Figures~\ref{fig_map}, \ref{fig_recall} and \ref{fig_rank} show that Qwen 2.5 7B outperforms Claude, when measured by MAP@k, recall@3 and median rank due to having better precision at low ranks. This is despite Claude having over an order of magnitude more parameters. On the other hand, Figures~\ref{fig_ndcg}, \ref{fig_recall} and \ref{fig_rank} show that Claude outperforms Qwen 2.5 7B when measured by NDCG, recall@10, and mean rank due to Claude predicting the true misconcepton in the top 10 more often. Overall, Claude and Qwen 2.5 7B perform comparably, despite Qwen being significantly smaller.

\subsection{Ablation of Stages}

\begin{figure}[htb!]
\centering
\includegraphics[width=\textwidth]{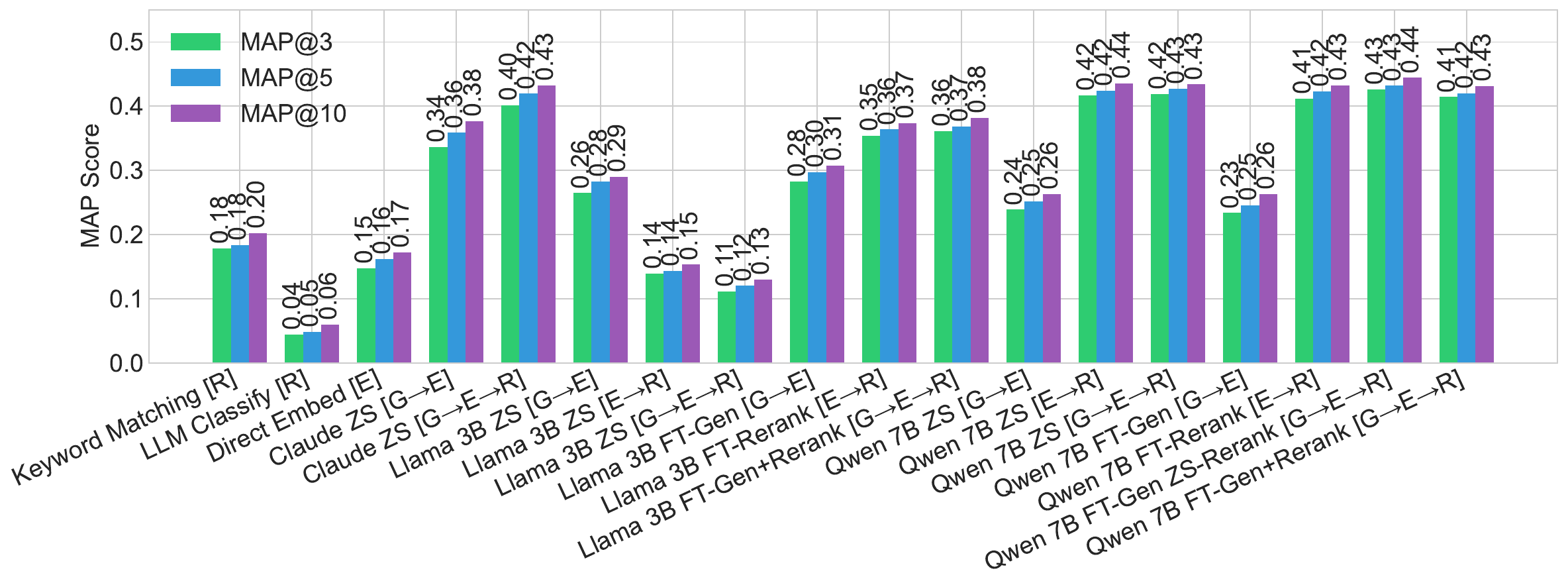}
\vspace{-0.9cm}
\caption{MAP@k. The [G $\rightarrow$ E $\rightarrow$ R] stands for [Generate $\rightarrow$ Embed $\rightarrow$ Re-rank] which signifies if the model is using the Generate stage, Embedding stage, and Re-rank stage.} \label{fig_map}
\end{figure}

\begin{figure}[htb!]
\centering
\includegraphics[width=\textwidth]{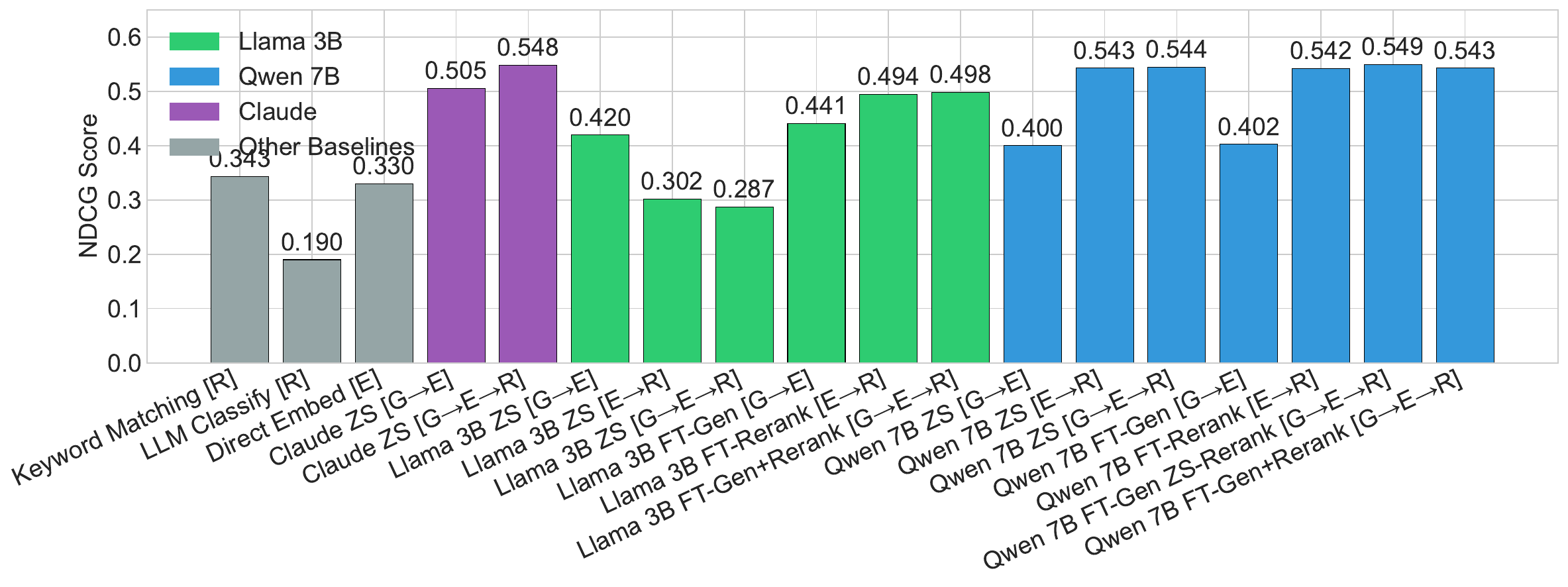}
\vspace{-0.9cm}
\caption{NDCG. Model notation is the same as defined in \Cref{fig_map}. 
}
\label{fig_ndcg}
\end{figure}

\begin{figure}[htb!]
\centering
\includegraphics[width=\textwidth]{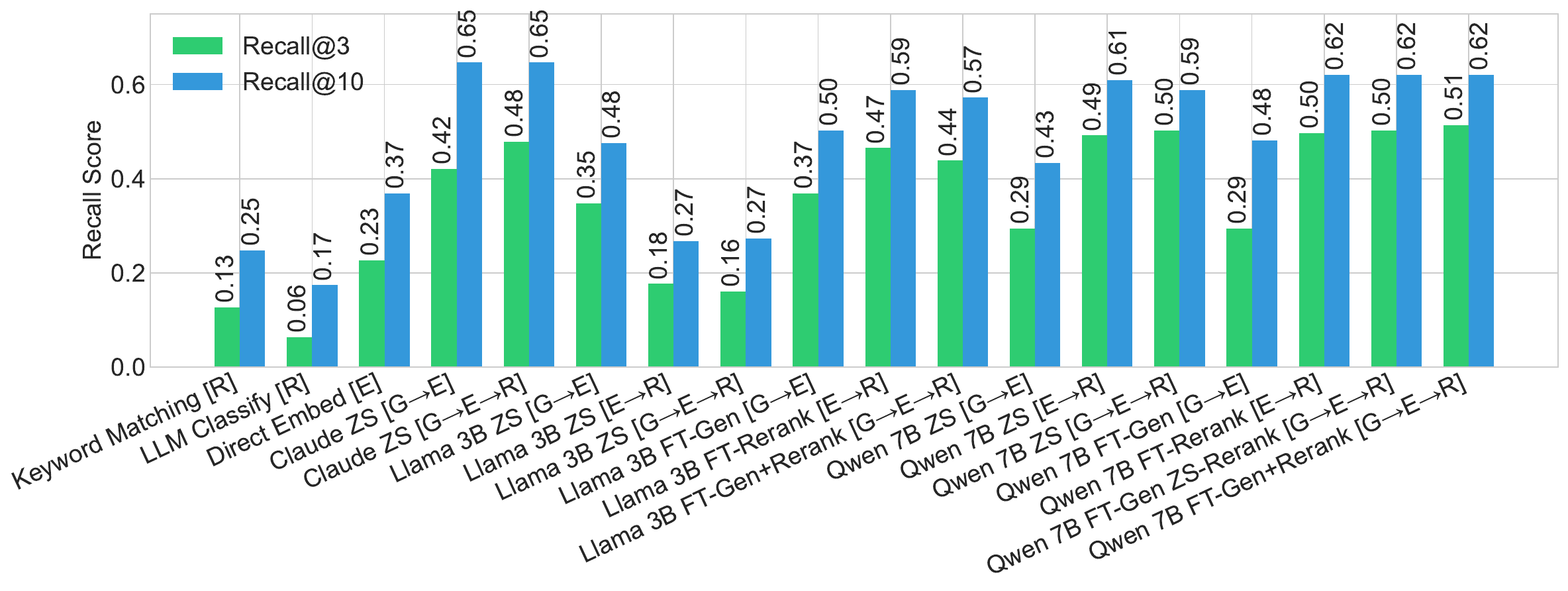}
\vspace{-0.9cm}
\caption{Recall. Model notation is the same as defined in \Cref{fig_map}.}
\label{fig_recall}
\end{figure}

\textit{Generative Stage:} The generative stage overall improves the model performance. Figures~\ref{fig_map} and \ref{fig_ndcg} show that for the fine-tuned Llama 3.2 3B LoRA model, zero-shot Qwen 2.5 7B, and fine-tuned Qwen 2.5 7B LoRA the MAP@k and NDCG performance is either equivalent or improved by the inclusion of the generative stage of the model. The zero-shot Llama 3.2 3B score is not informative due to the model having very poor reranking baseline ability. Considering the recall@3 and recall@10, in Figure~\ref{fig_recall}, the fine-tuned Llama 3.2 3B scored a lower rank when including the generative stage. This is most likely a weakness of the base not fully understanding the maths problem being considered. 
For Qwen 2.5 7B zero-shot including the generative stage has a minor improvement in recall@3 and minorly degrades recall@10. For the fine-tuned Qwen 2.5 7B LoRA then generative stage improves performance. 
Overall, the best performing model is the fine-tuned Qwen 2.5 7B LoRA model using the generative stage. 

\textit{Reranking:} From our experiments, reranking improves the final predicted misconceptions. Figures~\ref{fig_map}, \ref{fig_ndcg}, \ref{fig_recall} and \ref{fig_rank} demonstrate that for Claude Sonnet 4.5 including the reranking stage improves all metrics. This is because the reranking stage uses Claude's improved ability over the embedding model to understand the semantics of the generated misconception and rank the true misconception at a lower rank. Similarly, Qwen 2.5 7B also improves across all metrics by including the reranking stage, again showing that the LLM has better semantic understanding of the generated misconception. 
Llama 3.2 3B, being the smallest LLM considered, is an outlier to this result and its zero-shot reranking makes the performance worse across all metrics. On the other hand, the fine-tuned LoRA Llama 3.2 3B model improves across all metrics when including the reranking stage. 

\begin{figure}[htb!]
\centering
\includegraphics[width=\textwidth]{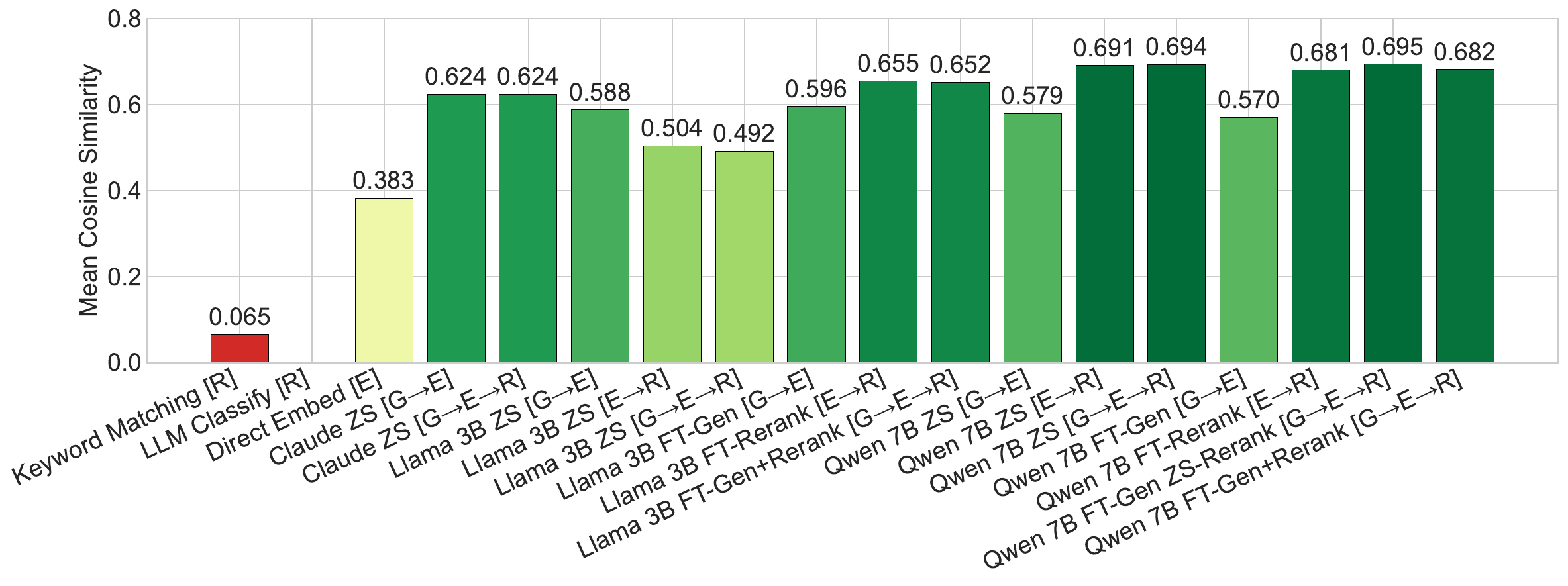}
\vspace{-0.9cm}
\caption{Cosine Similarity. Model notation is the same as defined in \Cref{fig_map}.}
\label{fig_cosinesim}
\end{figure}

\textit{Fine-tuning Open Source Models:}
Figures~\ref{fig_map}, \ref{fig_ndcg}, \ref{fig_cosinesim}, \ref{fig_rank} and \ref{fig_recall} all show that fine-tuning Llama 3.2 3B LoRA improves each possible model type across all metrics. This shows the benefit of fine-tuning small open source LLM models for specific tasks. Although the fine-tuned Llama model doesn't in general outperform the Claude model the performance drastically improves from fine-tuning and the model is significantly smaller. The Qwen 2.5 7B model has smaller improvements from fine-tuning the LoRA layers when compared to Llama, but this is a result of the base model having strong performance. Nevertheless, the fine-tuning of Qwen 2.5 7B does in general improve the model across each of the metrics and notably improves the NDCG score such that the model outperforms Claude.

\FloatBarrier

%% file: chapters/discussion.tex
\subsection{How Important is Base Model Selection?}
Our results show that base model choice significantly impacts misconception hypothesis generation quality. Models with stronger zero-shot reasoning capabilities, such as Claude Sonnet 4.5 and Qwen 2.5 7B, produce more accurate misconception descriptions than Llama 3.2 3B. This likely reflects differences in training data and alignment. Qwen 2.5 suite of models for example is known for strong math and coding performance. Models exposed to more mathematical and educational content may better recognize common student errors. Notably, Qwen 2.5 7B performs comparably to Claude Sonnet 4.5 despite being much smaller. This suggests that domain-relevant pre-training or post-training may matter more than raw parameter count for this task.

\subsection{Does Fine-tuning help Generation?}
Smaller open-source models often struggle to articulate misconceptions from dialogue alone. Prompt engineering can help but cannot reliably produce outputs in a desired style. To address this, we fine-tune the generator base model for both LLaMA 3.2 3B Instruct and Qwen 2.5 7B Instruct using Low-Rank Adaptation (LoRA). 
Table~\ref{tab:generation_comparison} shows the difference before and after fine-tuning on the held out 20\% test set. Zero-shot outputs tend to be verbose. These lengthy descriptions don't match the concise style of expert-authored labels. After fine-tuning, outputs typically match the format tutors use. We see higher cosine similarity between fine-tuned outputs and expert labels in the embedding space as compared to zero-shot misconception hypotheses for both open-source models, in \Cref{fig_cosinesim}. The fine-tuned model thus learns to produce descriptions that are stylistically closer to tutor annotations.
\input{tables/base_vs_finetune}

\subsection{Does Semantic Similarity Guarantee Mathematical Coherence?}
Embedding-based approaches demonstrate strong performance in matching misconception descriptions. However, semantic similarity does not guarantee mathematical equivalence. A model can identify misconceptions that are stylistically similar, i.e., sharing vocabulary or mathematical domain, but they may fundamentally differ in their underlying mathematical reasoning.

This distinction is critical because misconceptions that share vocabulary or domain, such as involving decimals, may require entirely different pedagogical interventions. A student who ``rounds decimals less than 1 to 0'' requires different instruction from a student who ``believes dividing by a decimal is the same as dividing by its reciprocal''. Table~\ref{tab:semantic_vs_mathematical} presents representative examples from our Claude Sonnet 4.5 evaluation where the predicted misconception matches the actual misconception stylistically but fails to capture the correct mathematical reasoning. These cases highlight systematic patterns where current embedding models succeed at surface-level linguistic matching but miss deeper mathematical structure. Future work should explore math-centric language models that understand mathematical structure as base models. While our approach achieves strong performance on standard metrics, in the future we also wish to explore better retrieval metrics that capture semantic similarity.
\input{tables/semantic_similarity_vs_mathematical_coherence}

\subsection{Does Reranking Improve Retrieval?}
While MiniLM-L6-v2 embeddings \cite{wang2020minilm} with cosine similarity provide good initial retrieval, they rely heavily on lexical overlap and word co-occurrence patterns. They struggle to capture subtle semantic distinctions between misconceptions. Reranking the top-k misconception hypotheses with LLMs can significantly improve rank-1 accuracy by understanding semantic relationships beyond surface-level word matching. Our re-rank approach, which re-ranks the top retrieved candidates, successfully promoted the correct misconception to rank 1 in 73 cases for Claude Sonnet 4.5 out of 229 reranked cases.
\input{tables/llm_reranking_improvements}
Table~\ref{tab:reranking_rank1_improvements} presents representative examples where re-ranking moved the correct misconception from rank 2 or lower to rank 1. These cases demonstrate a systematic pattern: the embedding model retrieves semantically close candidates but cannot distinguish between paraphrases or conceptually equivalent descriptions that use different vocabulary. These patterns suggest that future work could be to improve the embedding model to better understand the domain-specific semantic differences between mathematical misconception pairs. 

We also find that fine-tuning dedicated reranking models, creating dedicated misconception reranking models, improves performance. Further, fine-tuning open source models that are an order of magnitude smaller than large closed source models offers the best reranking performance. This suggests that on specific tasks such as understanding the semantics of misconceptions and reranking them, smaller specialised models are very performant.

%% file: tables/base_vs_finetune.tex
\begin{table*}[t]
\centering
\caption{Comparison of generated misconception descriptions between zero-shot and fine-tuned Qwen 2.5 7B. Fine-tuning produces more concise outputs that are stylistically closer to expert-authored labels.}
\vspace{-0.1cm}
\label{tab:generation_comparison}
\resizebox{\textwidth}{!}{%
\renewcommand{\arraystretch}{1.4}
\setlength{\tabcolsep}{12pt}
\begin{tabular}{p{2.5cm}p{5.5cm}p{4.5cm}p{4cm}}
\toprule
\textbf{Topic} & \textbf{Zero-Shot Generation} & \textbf{Fine-Tuned Generation} & \textbf{Actual Label} \\
\midrule
Mental multiplication by 99 
& The student mistakenly believes that multiplying by 100 and then adding the original number will give the correct result, when in fact, multiplying by 100 and subtracting the original number is the correct method. 
& Believes multiplying by 100 and adding the original number will give the same result as multiplying by 99 
& When performing mental multiplication, multiplies accurately by a multiple of 10 but then does not know how to adjust their answer \\
\addlinespace[0.3cm]
Understanding multiples 
& The student misconceives that a multiple must be obtained by multiplying a number by a whole number, not realizing that the result also needs to be a whole number. 
& Does not understand what a multiple is 
& Does not understand the term multiple \\
\bottomrule
\end{tabular}%
}
\vspace{-0.4cm}
\end{table*}

%% file: tables/semantic_similarity_vs_mathematical_coherence.tex
\begin{table*}[htb]
\centering
\caption{Examples where predicted misconceptions match stylistically and share domain/vocabulary, but differ mathematically, with explanations of the mathematical incoherence.}
\small
\setlength{\tabcolsep}{8pt}
\begin{tabular}{p{0.28\textwidth}p{0.28\textwidth}p{0.35\textwidth}}
\toprule
\textbf{Predicted Misconception} & \textbf{Actual Misconception} & \textbf{Mathematical Incoherence} \\
\midrule
Rounds decimals less than 1 to 0 when estimating & Believes that dividing by a decimal is the same as dividing by its reciprocal & Predicted is an estimation error; actual is fundamental misunderstanding of division operations with decimals \\
\addlinespace
Calculates percentage of new value instead of original value & Believes that they can reverse a percentage increase by decreasing the new value by the same percentage, and vice versa & Predicted uses wrong base value; actual concerns asymmetry of percentage changes \\
\bottomrule
\end{tabular}
\label{tab:semantic_vs_mathematical}
\vspace{-0.6cm}
\end{table*}

%% file: tables/llm_reranking_improvements.tex
\begin{table*}[t]
\centering
\caption{Examples where re-ranking successfully moved the correct misconception to rank 1. The baseline rank shows where embedding-based retrieval placed it; re-ranking promoted these to rank 1 by understanding semantic equivalence.}
\small
\setlength{\tabcolsep}{12pt}
\begin{tabular}{p{0.08\textwidth}p{0.36\textwidth}p{0.36\textwidth}}
\toprule
\textbf{Baseline Rank} & \textbf{Embedding Prediction} & \textbf{Actual Label (Re-ranked to Rank 1)} \\
\midrule
2 & Believes the mean is the median & When asked for the mean of a list of data, gives the median \\
\addlinespace
2 & Reverses the direction of a translation vector & When describing a translation, goes from the image to the original \\
\bottomrule
\end{tabular}
\label{tab:reranking_rank1_improvements}
\vspace{-0.2cm}
\end{table*}

%% file: chapters/appendix.tex
\section{Additional Metrics}
Another metric that we use to understand how the model is performing is the mean and median rank. These provide an overall view on the models ranking ability. The equations for mean and median rank are given by:
\begin{equation}
    \mathrm{Mean Rank} = \frac{1}{N} \sum^{N}_{i=1} r_{i}
    \label{eq:mean}
\end{equation}
and
\begin{equation}
    \mathrm{Median Rank} = \mathrm{median}(\{r_{1},...,r_{\abs{N}}\}),
    \label{eq:median}
\end{equation}
where $N$ is the number of data points, $r_{i}$ is the predicted rank of the true misconception for the $i^{\mathrm{th}}$ data point.

Finally, we also consider the cosine similarity of the generated misconception hypothesis with the true misconception text in embedding space. This provides a measure of stylistic similarity between the generated misconception and the ground-truth misconception label. This enables us to isolate the performance of Stage 1 from Stage 2 and 3. The equation we use for cosine similarity is:
\begin{equation}
    \mathrm{Cosine Similarity} (T, G) = \frac{T \cdot G}{\norm{T} \norm{G}},
    \label{eq:cs}
\end{equation}
where $T$ is the embedding vector from the true misconception text and $G$ is the embedding vector from the input from our model into the embedding layer.
\section{Results}
\begin{figure}[htb!]
\includegraphics[width=\textwidth]{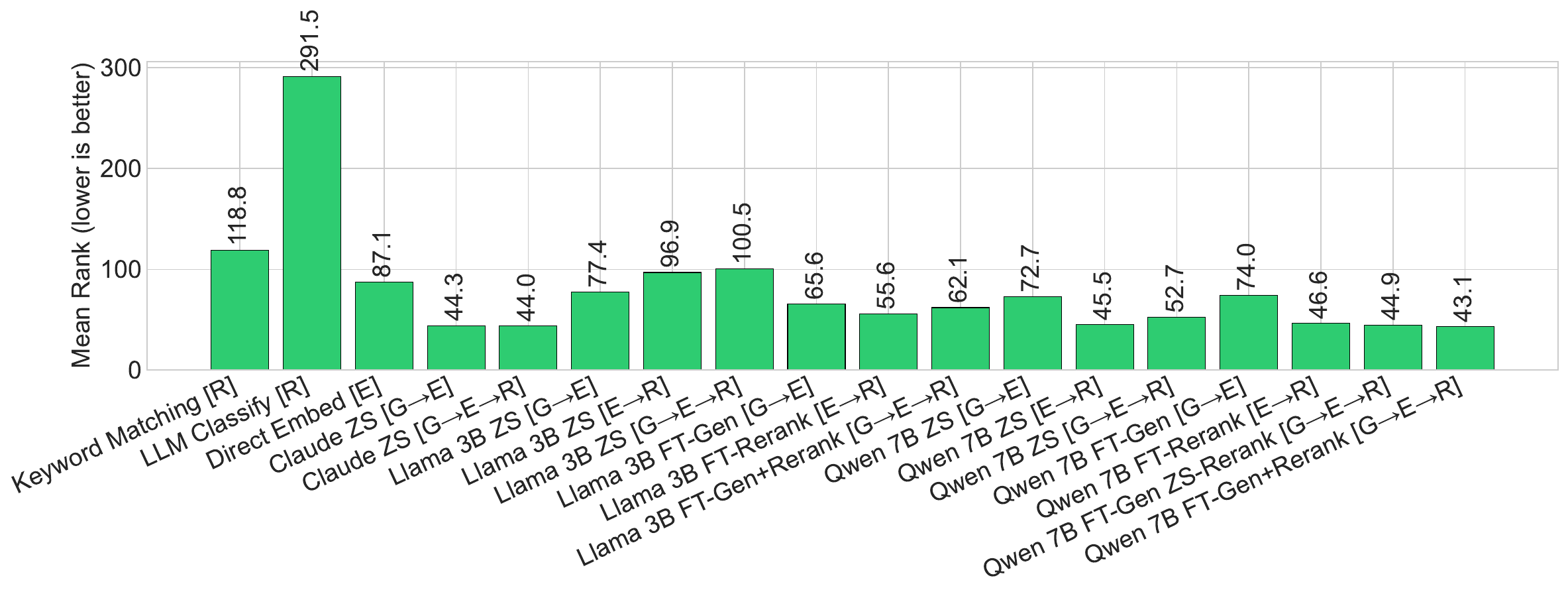}
\caption{Mean rank predicted by each model. Model notation is the same as defined in \Cref{fig_map}.} \label{fig_rank}
\end{figure}

\begin{figure}[htb!]
\includegraphics[width=\textwidth]{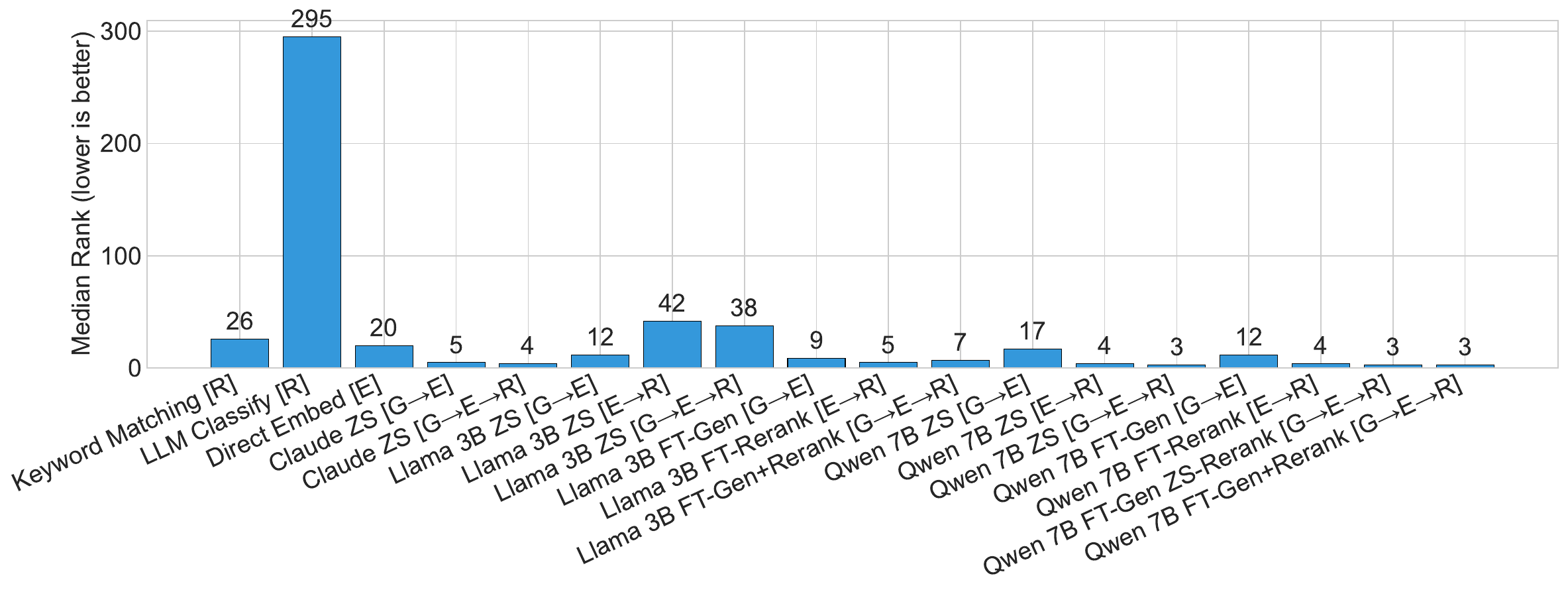}
\caption{Median rank predicted by each model. Model notation is the same as defined in \Cref{fig_map}.} \label{fig_rank_median}
\end{figure}

We show the mean and median rank of retrieved  misconceptions, when compared to tutor-annotations, in \Cref{fig_rank,fig_rank_median}.

\section{Additional Analysis of Model Outputs}
We expand on the results we provided in the Discussion section in the main paper. We show more examples of how fine-tuning improves the generative style to match that of the tutor-annotations in \Cref{tab:generation_comparison_appendix}. We also provide more cases where semantic similarity of the predicted misconceptions from our closed source model output does not necessarily lead to mathematical coherence when compared to the tutor labels in \Cref{tab:semantic_vs_mathematical_appendix}. Finally, we show cases where reranking significantly improves the top-1 retrieved misconception for our closed-source model, thereby showing the importance of this proposed stage in \Cref{tab:reranking_rank1_improvements_appendix}. \Cref{tab:rank_improvements} shows the distribution of rank improvements achieved through LLM reranking across 73 cases, demonstrating the model's ability to successfully promote the correct misconception from various baseline ranks to the top position.
\input{tables/base_vs_finetune_appendix}
\input{tables/semantic_similarity_vs_mathematical_coherence_appendix}
\input{tables/llm_reranking_improvements_appendix}
\input{tables/reranking_improvement_claude}
\section{Prompt Engineering for Misconception Generation and Re-ranking}
Effective prompt design proved critical for both misconception generation and re-ranking with closed-source LLMs. We systematically evaluated six prompting strategies for Claude Sonnet 4.5 on a sample of 20 conversations, measuring semantic similarity between generated predictions and ground-truth labels using MiniLM-L6-v2 \cite{wang2020minilm} embeddings. \Cref{tab:prompt_strategies} shows that naive prompts (e.g., ``identify the core mathematical misconception in 1-2 sentences'') produced verbose explanations averaging 56.8 words with a mean similarity of 0.550, performing poorly in embedding-based retrieval due to lexical dilution.

\input{tables/prompt_engineering}

Our best-performing strategy, ``With Examples'' achieved a mean similarity of 0.615 (12\% improvement over baseline) while reducing output length from 56.8 to 9.1 words (84\% reduction) by explicitly constraining output to 5-12 words and providing both positive examples (``Carries out operations from left to right regardless of priority order'') and negative anti-examples (``The student fails to understand that when performing operations...''). Critically, explicit length constraints alone were insufficient. ``Concise Label'' and ``Extract Key Concept'' both constrained length but achieved lower similarity (0.601 and 0.545) than ``With Examples,''. This demonstrates that negative anti-examples are essential for preventing the verbose explanatory style LLMs naturally produce. The final generation prompt includes the question, answer options, student selection, and dialogue, followed by instruction to ``Generate a concise misconception label matching these good examples'' with five representative labels, and ``Avoid verbose explanations like these bad examples'' with three anti-patterns.

For re-ranking, we developed a prompt emphasizing semantic reasoning over lexical matching. Given the generated prediction and top-k candidates from embedding retrieval, the model receives the instruction ``Rerank these by semantic similarity to the prediction. Focus on: (1) Core mathematical concept, (2) Type of error, (3) Underlying reasoning flaw,''. It outputs only the $k$ reordered numbers as a comma-separated list. These findings suggest that future work on fine-tuning models for misconception diagnosis should incorporate similar constraints and examples directly into the training objective, rather than relying solely on prompt engineering at inference time.

%% file: tables/base_vs_finetune_appendix.tex
\begin{table*}[htb]
\centering
\caption{Comparison of generated misconception descriptions between zero-shot and fine-tuned Qwen 2.5 7B. Fine-tuning produces more concise outputs that are stylistically closer to expert-authored labels.}
\label{tab:generation_comparison_appendix}
\resizebox{\textwidth}{!}{%
\renewcommand{\arraystretch}{1.4}
\setlength{\tabcolsep}{12pt}
\begin{tabular}{p{2.5cm}p{5.5cm}p{4.5cm}p{4cm}}
\toprule
\textbf{Topic} & \textbf{Zero-Shot Generation} & \textbf{Fine-Tuned Generation} & \textbf{Actual Label} \\
\midrule
Exponent rules ($16^0$) 
& The student misconceived the meaning of 0 as an exponent, thinking it results in the base number itself instead of 1. 
& Does not know that any number to the power of zero is 1 
& Forgotten that anything to the power of 0 equals 1 \\
\addlinespace[0.3cm]
Adding fractions 
& The student incorrectly added the numerators and denominators separately without finding a common denominator. 
& When adding fractions adds the numerators and denominators 
& Forgets to change the numerators when finding equivalent fractions \\
\addlinespace[0.3cm]
Algebraic simplification 
& The student misconceives that the variables $a$ and $b$ should be added together rather than multiplied. 
& Does not know how to multiply algebraic expressions 
& Thinks only like terms can be multiplied \\
\bottomrule
\end{tabular}%
}
\end{table*}

%% file: tables/semantic_similarity_vs_mathematical_coherence_appendix.tex
\begin{table*}[htp]
\centering
\caption{Examples where predicted misconceptions match stylistically but differ mathematically from actual misconceptions, with explanations of the mathematical incoherence.}
\small
\setlength{\tabcolsep}{8pt}
\begin{tabular}{p{0.28\textwidth}p{0.28\textwidth}p{0.35\textwidth}}
\toprule
\textbf{Predicted Misconception} & \textbf{Actual Misconception} & \textbf{Mathematical Incoherence} \\
\midrule
Subtracts one given angle from another instead of using angle sum & Thinks that all angles anywhere along a straight line add to 180 & Predicted is operational choice error; actual is overgeneralization of when 180° rule applies \\
\addlinespace
Averages middle two values instead of finding their mean & When finding the median from an even dataset does not understand we must find the midpoint of the two values in the middle & Predicted suggests student calculated mean; actual is that student doesn't understand median requires midpoint \\
\addlinespace
Adds or subtracts instead of using inverse operation to solve & Does not know how to interpret a fraction worded problem & Predicted is about solving equations with wrong operations; actual is about not understanding fraction word problems at all \\
\addlinespace
Believes gradient is the horizontal change rather than rise/run & Believes the gradient is where the line intersects the y-axis & Predicted confuses gradient with one component; actual confuses gradient with y-intercept \\
\bottomrule
\end{tabular}
\label{tab:semantic_vs_mathematical_appendix}
\end{table*}

%% file: tables/llm_reranking_improvements_appendix.tex
\begin{table*}[htb]
\centering
\caption{Examples where re-ranking successfully moved the correct misconception to rank 1. The baseline rank shows where embedding-based retrieval placed it; re-ranking promoted these to rank 1 by understanding semantic equivalence beyond lexical matching.}
\small
\setlength{\tabcolsep}{12pt}
\begin{tabular}{p{0.08\textwidth}p{0.36\textwidth}p{0.36\textwidth}}
\toprule
\textbf{Baseline Rank} & \textbf{Embedding Prediction} & \textbf{Actual Label (Re-ranked to Rank 1)} \\
\midrule
2 & Writes the base number repeatedly instead of multiplying it & Writes the digit 3 times instead of cubing \\
\addlinespace
2 & Ignores carried digit when performing column addition & When two digits sum to 10 or more during an addition problem, does not add one to the preceding digit \\
\addlinespace
2 & Converts mixed number to improper fraction but doesn't add whole number & Thinks you add the number of wholes to the numerator when converting a mixed number to an improper fraction \\
\addlinespace
2 & Believes the number with more decimal places is closer to zero & Does not know how to compare decimal numbers \\
\addlinespace
6 & Does not recognize algebraically equivalent expressions as equal & Does not think a factorised expression is equivalent to its multiplied out form \\
\addlinespace
7 & Subtracts the absolute value when subtracting a negative number & Does not realise that subtracting a negative is the same as adding it's additive inverse \\
\addlinespace
6 & Confuses place value with digit value by one place & Confuses the ten thousands and thousands place value columns \\
\bottomrule
\end{tabular}
\label{tab:reranking_rank1_improvements_appendix}
\end{table*}

%% file: tables/reranking_improvement_claude.tex
\begin{table}[h]
\centering
\begin{tabular}{ccc}
\toprule
\textbf{Baseline Rank} & \textbf{Final Rank} & \textbf{Number of Cases} \\
\midrule
2 & 1 & 24 \\
3 & 1 & 7 \\
4 & 1 & 10 \\
5 & 1 & 8 \\
6 & 1 & 4 \\
7 & 1 & 10 \\
8 & 1 & 2 \\
9 & 1 & 7 \\
10 & 1 & 1 \\
\midrule
\textbf{Total} & & \textbf{73} \\
\bottomrule
\end{tabular}
\caption{Rank improvements achieved by LLM reranking (Claude Sonnet 4.5)}
\label{tab:rank_improvements}
\end{table}

%% file: tables/prompt_engineering.tex
\begin{table}[t]
\centering
\small
\begin{tabular}{lcc}
\toprule
\textbf{Prompt Strategy} & \textbf{Mean Similarity} & \textbf{Average Words} \\
\midrule
With Examples (best) & \textbf{0.615} & \textbf{9.1} \\
Concise Label & 0.601 & 9.4 \\
Few-Shot & 0.575 & 11.5 \\
Original (verbose) & 0.550 & 56.8 \\
Structured & 0.546 & 12.5 \\
Extract Key Concept & 0.545 & 9.2 \\
\bottomrule
\end{tabular}
\caption{Prompt engineering results on 20 test conversations. ``With Examples'' achieved a 12\% improvement in semantic similarity and 84\% reduction in word count compared to the original verbose prompt.}
\label{tab:prompt_strategies}
\end{table}

%% file: references.bib
@article{brown1978diagnostic,
  title={Diagnostic models for procedural bugs in basic mathematical skills},
  author={Brown, John Seely and Burton, Richard R},
  journal={Cognitive science},
  volume={2},
  number={2},
  pages={155--192},
  year={1978},
  publisher={Elsevier}
}

@book{vanlehn1990mind,
  title={Mind bugs: The origins of procedural misconceptions},
  author={VanLehn, Kurt},
  year={1990},
  publisher={MIT press}
}

@inproceedings{feldman2018automatic,
  title={Automatic diagnosis of students' misconceptions in k-8 mathematics},
  author={Feldman, Molly Q and Cho, Ji Yong and Ong, Monica and Gulwani, Sumit and Popovi{\'c}, Zoran and Andersen, Erik},
  booktitle={Proceedings of the 2018 CHI Conference on Human Factors in Computing Systems},
  pages={1--12},
  year={2018}
}

@article{ross2024toward,
  title={Toward In-Context Teaching: Adapting Examples to Students' Misconceptions},
  author={Ross, Alexis and Andreas, Jacob},
  journal={arXiv preprint arXiv:2405.04495},
  year={2024}
}

@inproceedings{sonkar2024malalgoqa,
  title={Malalgoqa: Pedagogical evaluation of counterfactual reasoning in large language models and implications for ai in education},
  author={Sonkar, Shashank and Liu, Naiming and Le, MyCo and Baraniuk, Richard},
  booktitle={Findings of the Association for Computational Linguistics: EMNLP 2024},
  pages={15554--15567},
  year={2024}
}

@inproceedings{scarlatos2025exploring,
  title={Exploring knowledge tracing in tutor-student dialogues using llms},
  author={Scarlatos, Alexander and Baker, Ryan S and Lan, Andrew},
  booktitle={Proceedings of the 15th International Learning Analytics and Knowledge Conference},
  pages={249--259},
  year={2025}
}

@article{ross2025learning,
  title={Learning to make mistakes: Modeling incorrect student thinking and key errors},
  author={Ross, Alexis and Andreas, Jacob},
  journal={arXiv preprint arXiv:2510.11502},
  year={2025}
}

@misc{eedi-mining-misconceptions-in-mathematics,
    author = {Jules King and L Burleigh and Simon Woodhead and Panagiota Kon and Perpetual Baffour and Scott Crossley and Walter Reade and Maggie Demkin},
    title = { Eedi - Mining Misconceptions in Mathematics},
    year = {2024},
    howpublished = {\url{https://kaggle.com/competitions/eedi-mining-misconceptions-in-mathematics}},
    note = {Kaggle}
}

@article{reimers2019sentence,
  title={Sentence-bert: Sentence embeddings using siamese bert-networks},
  author={Reimers, Nils and Gurevych, Iryna},
  journal={arXiv preprint arXiv:1908.10084},
  year={2019}
}

@inproceedings{karpukhin2020dense,
  title={Dense Passage Retrieval for Open-Domain Question Answering.},
  author={Karpukhin, Vladimir and Oguz, Barlas and Min, Sewon and Lewis, Patrick SH and Wu, Ledell and Edunov, Sergey and Chen, Danqi and Yih, Wen-tau},
  booktitle={EMNLP (1)},
  pages={6769--6781},
  year={2020}
}

@article{wang2024multilingual,
  title={Multilingual e5 text embeddings: A technical report},
  author={Wang, Liang and Yang, Nan and Huang, Xiaolong and Yang, Linjun and Majumder, Rangan and Wei, Furu},
  journal={arXiv preprint arXiv:2402.05672},
  year={2024}
}

@inproceedings{su2023one,
  title={One embedder, any task: Instruction-finetuned text embeddings},
  author={Su, Hongjin and Shi, Weijia and Kasai, Jungo and Wang, Yizhong and Hu, Yushi and Ostendorf, Mari and Yih, Wen-tau and Smith, Noah A and Zettlemoyer, Luke and Yu, Tao},
  booktitle={Findings of the Association for Computational Linguistics: ACL 2023},
  pages={1102--1121},
  year={2023}
}

@inproceedings{muennighoff2023mteb,
  title={Mteb: Massive text embedding benchmark},
  author={Muennighoff, Niklas and Tazi, Nouamane and Magne, Lo{\"\i}c and Reimers, Nils},
  booktitle={Proceedings of the 17th Conference of the European Chapter of the Association for Computational Linguistics},
  pages={2014--2037},
  year={2023}
}

@article{hu2022lora,
  title={Lora: Low-rank adaptation of large language models.},
  author={Hu, Edward J and Shen, Yelong and Wallis, Phillip and Allen-Zhu, Zeyuan and Li, Yuanzhi and Wang, Shean and Wang, Lu and Chen, Weizhu and others},
  journal={ICLR},
  volume={1},
  number={2},
  pages={3},
  year={2022}
}

@article{wang2020minilm,
  title={Minilm: Deep self-attention distillation for task-agnostic compression of pre-trained transformers},
  author={Wang, Wenhui and Wei, Furu and Dong, Li and Bao, Hangbo and Yang, Nan and Zhou, Ming},
  journal={Advances in neural information processing systems},
  volume={33},
  pages={5776--5788},
  year={2020}
}

@article{baral2025drawedumath,
  title={DrawEduMath: Evaluating Vision Language Models with Expert-Annotated Students' Hand-Drawn Math Images},
  author={Baral, Sami and Lucy, Li and Knight, Ryan and Ng, Alice and Soldaini, Luca and Heffernan, Neil T and Lo, Kyle},
  journal={arXiv preprint arXiv:2501.14877},
  year={2025}
}

@inproceedings{AKT,
author = {Ghosh, Aritra and Heffernan, Neil and Lan, Andrew S.},
title = {Context-Aware Attentive Knowledge Tracing},
year = {2020},
isbn = {9781450379984},
publisher = {Association for Computing Machinery},
address = {New York, NY, USA},
url = {https://doi.org/10.1145/3394486.3403282},
doi = {10.1145/3394486.3403282},
booktitle = {Proceedings of the 26th ACM SIGKDD International Conference on Knowledge Discovery \& Data Mining},
pages = {2330–2339},
numpages = {10},
location = {Virtual Event, CA, USA},
series = {KDD '20}
}

@article{MIRAGE,
  title={MiRAGE: Misconception Detection with Retrieval-Guided Multi-Stage Reasoning and Ensemble Fusion},
  author={Cuong Van Duc and Thai Tran Quoc and Minh Nguyen Dinh Tuan and Tam Vu Duc and Son Nguyen Van and Hanh Nguyen Thi},
  journal={ArXiv},
  year={2025},
  volume={abs/2511.01182},
  url={https://api.semanticscholar.org/CorpusID:282739399}
}

@article{sparck1972statistical,
  title={A statistical interpretation of term specificity and its application in retrieval},
  author={Sparck Jones, Karen},
  journal={Journal of documentation},
  volume={28},
  number={1},
  pages={11--21},
  year={1972},
  publisher={MCB UP Ltd}
}

@article{qwen2025qwen25technicalreport,
  title={Qwen2 Technical Report},
  author={Yang, An and Yang, Baosong and Hui, Binyuan and Zheng, Bo and Yu, Bowen and Zhou, Chang and Li, Chengpeng and Li, Chengyuan and Liu, Dayiheng and Huang, Fei and others},
  journal={CoRR},
  year={2024}
}

@misc{grattafiori2024llama3herd,
  title        = {The Llama 3 Herd of Models},
  author       = {Grattafiori, Aaron and Dubey, Abhimanyu and Jauhri, Akshita and others},
  year={2024},
      eprint={2407.21783},
      archivePrefix={arXiv},
      primaryClass={cs.AI},
      url={https://arxiv.org/abs/2407.21783}, 
}

@misc{anthropic2025claudeSonnet45SystemCard,
  author       = {Anthropic},
  title        = {Claude Sonnet 4.5 System Card},
  year         = {2025},
  month        = oct,
  url          = {https://www.anthropic.com/claude-sonnet-4-5-system-card},
  note         = {Accessed 2026-01-27}
}

@article{wang2020diagnostic,  
    title={Diagnostic questions: The neurips 2020 education challenge},
    author={Wang, Zichao and Lamb, Angus and Saveliev, Evgeny and Cameron, Pashmina and Zaykov, Yordan and Hern{\'a}ndez-Lobato, Jos{\'e} Miguel and Turner, Richard E and Baraniuk, Richard G and Barton, Craig and Jones, Simon Peyton and Woodhead, Simon and Zhang, Cheng},
    journal={arXiv preprint arXiv:2007.12061},  
    year={2020}
}
